\documentclass[12pt, draftclsnofoot, onecolumn]{IEEEtran}
\usepackage{bm}
\usepackage{bbm}
\usepackage{mathrsfs}
\usepackage{textcomp}
\usepackage{epsfig}
\usepackage{indentfirst}
\usepackage{amsmath}
\usepackage{amssymb}
\usepackage{array}
\usepackage{multirow}
\usepackage{graphicx}
\usepackage{subfigure}
\usepackage{multicol}

\usepackage{algorithm}
\usepackage{algpseudocode}

\usepackage{graphicx}
\usepackage{epstopdf}
\usepackage{color}

\usepackage{amsthm}

\usepackage{xcolor}
\usepackage{soul}

\theoremstyle{plain}

\theoremstyle{definition}

\theoremstyle{remark}

\usepackage{cite}

\begin{document}
\title{Decentralized Computation Offloading for Multi-User Mobile Edge Computing: A Deep Reinforcement Learning Approach}
\author{{Zhao Chen,~\IEEEmembership{Member,~IEEE} and Xiaodong Wang,~\IEEEmembership{Fellow,~IEEE}}
\thanks{Z.~Chen was with the Department of Electrical Engineering, Columbia University, New York, NY 10027, USA. He is now with Amazon Canada, Vancouver, BC V6B 0M3, Canada (e-mail: zhaochen@ieee.org).}
\thanks{X. Wang is with the Department of Electrical Engineering, Columbia University, New York, NY 10027, USA (e-mail: wangx@ee.columbia.edu).}}
\maketitle

\vspace{-1em}
\begin{abstract}
Mobile edge computing (MEC) emerges recently as a promising solution to relieve resource-limited mobile devices from computation-intensive tasks, which enables devices to offload workloads to nearby MEC servers and improve the quality of computation experience. 
Nevertheless, by considering an MEC system consisting of multiple mobile users with stochastic task arrivals and wireless channels in this paper, the design of computation offloading policies is challenging to minimize the long-term average computation cost in terms of power consumption and buffering delay. A deep reinforcement learning (DRL) based decentralized dynamic computation offloading strategy is investigated to build a scalable MEC system with limited feedback.
Specifically, a continuous action space based DRL approach named deep deterministic policy gradient (DDPG) is adopted to learn efficient computation offloading policies independently at each mobile user.
Thus, powers of both local execution and task offloading can be adaptively allocated by the learned policies from each user's local observation of the MEC system. 
Numerical results are illustrated to demonstrate that efficient policies can be learned at each user, and performance of the proposed DDPG based decentralized strategy outperforms the conventional deep Q-network (DQN) based discrete power control strategy and some other greedy strategies with reduced computation cost. 
Besides, the power-delay tradeoff is also analyzed for both the DDPG based and DQN based strategies. 
\end{abstract}


\section{Introduction}

As the popularity of smart mobile devices in the coming 5G era, mobile applications, especially for computation-intensive tasks such as online 3D gaming, face recognition and location-based augmented or virtual reality (AR/VR), have been greatly affected by the limited on-device computation capability~\cite{zhang2013toward}.
Meanwhile, for the large number of low-power and resource-constrained wireless terminals serving in the emerging Internet of Things (IoT)~\cite{sun2016edgeiot} and Intelligent Transport Systems (ITS)~\cite{zhang2017mobile}, a huge amount of sensory data also needs to be pre-processed and analyzed. 
As a result, to meet the quality of experience (QoE) of these mobile applications, the technology of mobile edge computing (MEC)~\cite{satyanarayanan2017emergence} has been proposed as a promising solution to bridge the gap between the limited resources on mobile devices and the ever-increasing demand of computation requested by mobile applications. 

Instead of the remote public clouds in conventional cloud computing systems such as Amazon Web Services and Microsoft Azure, MEC enhances the radio access networks (RANs), which is in close proximity to mobile users, with computing capability~\cite{shi2016edge,mach19mobile,mao2017survey}. It enables mobile devices to offload computation workloads to the MEC server associated with a base station (BS), and thus improves the QoE of mobile applications with considerably reduced latency and power consumption.
Many researcher has been attracted from both the industry~\cite{patel2014mobile} and academia~\cite{kumar2010cloud,kosta2012thinkair,huang2012dynamic}. 
Nevertheless, computation offloading highly depends on the efficiency of wireless data transmission, which requires MEC systems to manage radio resources along with computation resources and complete computation tasks efficiently. 

In order to achieve higher energy efficiency or better computation experience, computation offloading strategies for MEC have been widely investigated in the literature recently. 
For short-term optimization over quasi-static channels, some algorithms have been studied in~\cite{chen2018task,chen2015decentralized,du2018computation,guo2018mobile,zhang2018energy,bi2018computation,ding2018impact}.
In \cite{chen2018task}, optimal joint offloading selection and radio resource allocation for mobile task offloading was studied to minimize the overall execution time.
For decentralized algorithms with reduced overhead, a game-theoretic 
computation offloading scheme was constructed in~\cite{chen2015decentralized}.
Moreover, with the dynamic voltage and frequency (DVFS) techniques, CPU-cycle frequency was flexibly controlled with other features in~\cite{du2018computation,guo2018mobile}, where the system cost, defined as weighted sum of energy consumption and execution time, has been reduced. 
Besides, energy-latency tradeoff has been discussed in~\cite{zhang2018energy} with jointly optimized communication and computation resource allocation under the limited energy and sensitive latency. 
Also, it has been shown the performance of MEC can be further improved with adopting some other emerging technologies such as wireless power transfer~\cite{bi2018computation} and non-orthogonal multiple access (NOMA)~\cite{ding2018impact}. 

To cope with stochastic task arrivals and time-varying wireless channels, strategies for dynamic joint control of radio and computation resources in MEC systems become even challenging~\cite{zhang2013energy,kwak2015dream,sardellitti2015joint,mao2016dynamic,mao2017stochastic,lyu2017optimal,chen2018multi,liu2016delay,hong2016qoe,xu2017online}.
In~\cite{zhang2013energy}, a threshold-based dynamic computation offloading policy was proposed to minimize energy consumption under stochastic wireless channels. 
For low-complexity online algorithms, Lyapunov optimization has been widely adopted. 
In~\cite{kwak2015dream}, dynamic policies for offloading decision, clock speed and network interface control were considered 
to minimize energy consumption with given delay constraints. 
Joint optimization of multiple-input multiple-output (MIMO) beamforming and computational resource allocation for a multi-cell MEC system is designed in~\cite{sardellitti2015joint}. 
Additionally, an energy harvesting enabled green MEC system is studied in~\cite{mao2016dynamic}, where the delay cost addressing both the execution delay and task failure is minimized. 
For multiple user scenarios, power-delay tradeoff~\cite{mao2017stochastic}, network utility maximization balancing throughput and fairness with reduced feedback~\cite{lyu2017optimal}, and stochastic admission control and scheduling for multi-user multi-task computation offloading~\cite{chen2018multi} were discussed, respectively. 
On the other hand, Markov decision process (MDP) can be also applied to the analysis and design of dynamic control of MEC systems~\cite{liu2016delay,hong2016qoe}. 
Furthermore, it was shown in~\cite{xu2017online} and \cite{dinh2018distributed} that an optimal dynamic computation offloading policy can be learned by the emerging reinforcement learning (RL) based algorithm without any prior knowledge of the MEC system. 

Conventional RL algorithms cannot scale well as the number of agents increases, since the explosion of state space will make traditional tabular methods infeasible~\cite{sutton1998reinforcement}. 
Nevertheless, by exploiting deep neural networks (DNNs) for function approximation, deep reinforcement learning (DRL) has been demonstrated to efficiently approximate Q-values of RL~\cite{mnih2015human}. 
There have been some attempts to adopt DRL in the design of online resource allocation and scheduling in wireless networks~\cite{alqerm2017energy,he2018green,nasir2018deep}, especially for some recent works targeting computation offloading in MEC~\cite{chen2018optimized,huang2018deep,min2017learning,li2018deep}.
Specifically, in~\cite{li2018deep}, system sum cost combining execution delay and energy consumption of a multi-user MEC system is minimized by optimal offloading decision and computational resource allocation. Similarly, the authors in~\cite{huang2018deep} considered an online offloading algorithm to maximize the weighted sum computation rate in a wireless powered MEC system.
In~\cite{min2017learning}, a DRL based computation offloading
strategy of an IoT device is learned to choose one MEC server to offload and determine the offloading rate.
Besides, double deep Q-network (DQN) based strategic computation offloading algorithm was proposed in~\cite{chen2018optimized}, where an mobile device learned the optimal task offloading and energy allocation to maximize the long-term utility based on the task queue state, the energy queue state as well as the channel qualities. 
In the existing works, there have been only strategies focusing on centralized DRL based algorithms for optimal computation offloading in MEC systems, and the design of decentralized DRL based algorithms for dynamic task offloading control of a multi-user MEC system still remains unknown. 

In this paper, we consider a general MEC system consisting of one base station (BS) with one attached MEC server and multiple mobile users, where tasks arrive stochastically and channel condition is time-varying for each user.
Without any prior knowledge of network statistics of the MEC system,
a dynamic computation offloading policy will be learned independently at each mobile user based on local observations of the MEC system. 
Moreover, different from other DRL based policies in existing works making decisions in discrete action spaces, we adopt a continuous action space based algorithm named deep deterministic policy gradient (DDPG) to derive better power control of local execution and task offloading.
Specifically, major contributions of this paper can be summarized as follows:
\begin{itemize}
    \item A multi-user MIMO based MEC system is considered, where each mobile user with stochastic task arrivals and time-varying wireless channels attempts to independently learn dynamic computation offloading policies from scratch to minimize long-term average computation cost in terms of power consumption and task buffering delay. 
    \item By adopting DDPG, a DRL framework for decentralized dynamic computation offloading has been designed, which enables each mobile user to leverage only local observations of the MEC system to gradually learn efficient policies for dynamic power allocation of both local execution and computation offloading in a continuous domain. 
    \item Numerical simulations are performed to illustrate performance of the policy learned from the DDPG based decentralized strategy and analyze the power-delay tradeoff for each user. Superiority of the continuous power control based DDPG over the discrete control based DQN and some other greedy strategies is also demonstrated. 
\end{itemize}


The rest of this paper is organized as follows. In Section~\ref{sec.pre}, some preliminaries on DRL are introduced. 
Then, system model for the dynamic computation offloading of the MEC system is presented in Section~\ref{sec.MEC_Model}. 
Design of the decentralized DRL based dynamic computation offloading algorithm is proposed in Section~\ref{sec.DRL_design}. 
Numerical results will be illustrated in Section~\ref{sec.simulation}. Finally, Section~\ref{sec.conculusion} concludes this paper.

\section{Preliminaries on Deep Reinforcement Learning}\label{sec.pre}

In this section, we will firstly give an overview of MDP and RL~\cite{sutton1998reinforcement}, and then introduce some basics of the emerging DRL technology~\cite{mnih2015human}. Finally, recent extension of DRL on continuous action space based algorithm, i.e., DDPG~\cite{lillicrap2016continuous}, is presentedX
\subsection{MDP}
An MDP consists of an agent and an environment $E$, a set of possible states $\mathcal{S}$, a set of available actions $\mathcal{A}$, and a reward function $r : \mathcal{S} \times \mathcal{A} \rightarrow \mathcal{R}$, where the agent continually learns and makes decisions from the interaction with the environment in discrete time steps. 
In each time step $t$, the agent observes current state of the environment as $s_t \in \mathcal{S}$, and chooses and executes a action $a_t \in \mathcal{A}$ according to a policy $\pi$. After that, the agent will receive a scalar reward $r_t = r(s_t, a_t) \in \mathcal{R} \subseteq \mathbb{R}$ from the environment $E$ and find itself in the next state $s_{t+1} \in \mathcal{S}$ according to the transition probability of the environment $p(s_{t+1}|s_t,a_t)$. 
Thus, the dynamics of the environment $E$ is determined by the transition probability as response to the action taken by the agent in the current state, while the goal of the agent is to find the optimal policy that maximizes the long-term expected discounted reward it receives, i.e., 
\begin{align}
R_t = \sum_{i=t}^T \gamma^{i-t} r(s_i, a_i), 
\end{align} 
where $T \rightarrow \infty$ is the total number of time steps taken and $\gamma \in [0,1]$ is the discounting factor. 
It is worth noting that a policy $\pi$ is generally stochastic, which maps the current state to a probability distribution over the actions, i.e., $\pi : \mathcal{S} \rightarrow \mathcal{P}(\mathcal{A})$.
Under the policy $\pi$, the expected discounted return starting from state $s_t$ is defined as the value function 
\begin{align}
V^{\pi}(s_t,a_t) = \mathbb{E}_{s_{i>t} \sim E, a_{i \geq t} \sim \pi}\left[R_t | s_t\right],
\end{align}
while the state-action function is the expected discounted return after taking an action $a_t$, i.e., 
\begin{align}\label{eq.expected_return}
Q^{\pi}(s_t,a_t) = \mathbb{E}_{s_{i>t} \sim E, a_{i>t} \sim \pi}\left[R_t | s_t, a_t\right].
\end{align}
A fundamental property which is frequently used in MDP called Bellman equation represents the recursive relationship of the value function and the state-action function, respectively,
\begin{align}
V^{\pi}(s_t) & = \mathbb{E}_{s_{t+1} \sim E, a_t \sim \pi}\left[r(s_t,a_t)
+ \gamma V^{\pi}(s_{t+1}) \right], \\
Q^{\pi}(s_t,a_t) & = \mathbb{E}_{s_{t+1} \sim E}\left[r(s_t,a_t)
+ \gamma \mathbb{E}_{a_{t+1} \sim \pi} \left[ Q^{\pi}(s_{t+1},a_{t+1}) \right]\right].
\end{align}
Moreover, under the optimal policy $\pi^*$,
the Bellman optimality equation for the value function can be written by
\begin{align}\label{eq.opt_bellman_value}
V^{*}(s_t) = \max_{a_{t} \in \mathcal{A}} \mathbb{E}_{s_{t+1} \sim E}\left[r(s_t,a_t) + \gamma V^{*}(s_{t+1}) \right].
\end{align}
Based on the assumption of perfect model of the environment in MDP, dynamic programming (DP) algorithms like value iteration can be app∫lied to obtain the optimal value function of any state $s \in \mathcal{S}$ under the optimal policy $\pi^*$, i.e., 
\begin{align}
V_{k+1}(s) = \max_{a \in \mathcal{A}} \sum_{s^\prime} p(s^\prime | s, a) 
\left[ r(s, a) + \gamma V_{k}(s^\prime) \right],
\end{align}
where $k$ is the index for value iteration. Once the optimal value function $V^*(s)$ is obtained, the optimal state-value function can be derived by $Q^*(s,a) = \mathbb{E}_{s^\prime \sim E}\left[r(s,a)
+ \gamma V^*(s^\prime) \right]$. 
Then, it can be found that the optimal policy $\pi^*$ chooses the optimal action greedily in state $s$ as follows~\footnote{Note that this gives an special case for deterministic policies, which can be readily extended to stochastic policies. Specifically, the value iteration in \eqref{eq.opt_bellman_value} still holds for stochastic policies. If there are ties for different actions that maximized the value function, each maximizing action can be given a portion of probability to be selected, while other actions is selected with zero probability.},
\begin{align}\label{eq.opt_policy}
\pi^*(s) = \arg \max_{a \in \mathcal{A}} Q^*(s, a).
\end{align}


\subsection{RL}
Unlink MDP, RL algorithms attempt to derive optimal policies without an explicit model of the environment's dynamics. In this case, the underlying transmission probability $p(s_{t+1}|s_t,a_t)$ is unknown and even non-stationary. Thus, the RL agent will learn from actual interactions with the environment and adapting its behavior upon experiencing the outcome of its actions, so as to maximize the expected discounted rewards.
To this end, as a combination of Monte Carlo methods and DP, temporal-difference (TD) method arises to learn directly from raw experience.
Note that similar to \eqref{eq.opt_bellman_value}, the Bellman optimality equation for the state-action function is 
\begin{align}
Q^{*}(s_t,a_t) = \mathbb{E}_{s_{t+1} \sim E}\left[r(s_t,a_t)
+ \gamma \max_{a_{t+1}} Q^{*}(s_{t+1},a_{t+1}) \right],
\end{align}
from which we can update the state-action function using the agent's experience tuple $(s_t,a_t,r_t,s_{t+1})$ and other learned estimates at each time step $t$ as follows,
\begin{align}\label{eq.q_learning}
Q(s_t, a_t) \leftarrow Q(s_t,a_t) + \alpha \left[ r(s_t,a_t) + \gamma\max_{a_{t+1}} Q(s_{t+1}, a_{t+1}) - Q(s_t,a_t)\right],
\end{align}
where $\alpha$ is the learning rate and the value $\delta_t = r(s_t,a_t) + \gamma\max_{a_{t+1}} Q(s_{t+1}, a_{t+1}) - Q(s_t,a_t)$ is the TD error.
The algorithm in~\eqref{eq.q_learning} is the well-known Q-learning~\cite{watkins1992qlearning}, from which the state-action function is also known as Q-value.
It is worth noting that Q-learning is off-policy, since it directly approximates the optimal Q-value and the transitions experienced by the agent are independent of the policy being learned.
Besides, it can be proved that Q-learning algorithm converges with probability one~\cite{sutton1998reinforcement}. With the estimated optimal state-action function, the optimal policy $\pi^*$ can be easily obtained from~\eqref{eq.opt_policy}. 

\begin{figure}
\centering
\includegraphics[height = 8cm]{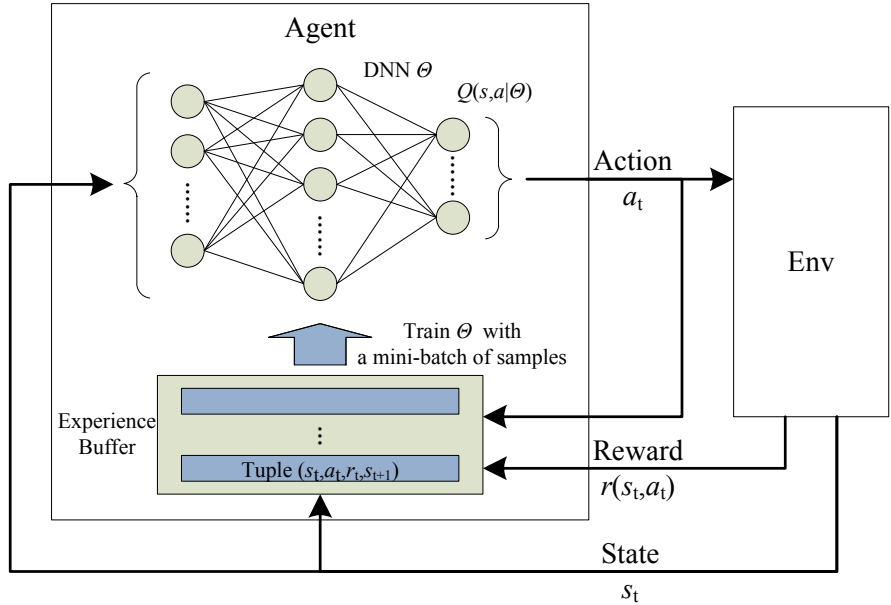}
\caption{Diagram of DQN.}
\label{fig.dqn}
\end{figure}

\subsection{DRL}
Thanks to the powerful function approximation properties of DNNs, DRL algorithms~\cite{mnih2015human} can learn low-dimensional representations for RL problems, which addresses the curse of dimensionality efficiently.
As illustrated in Fig.~\ref{fig.dqn}, the recent DQN technology~\cite{mnih2015human} successfully takes advantage of a DNN parameterized by $\theta$ to approximate the Q-values $Q(s,a)$.
In order to resolve the problem of instability of using function approximation in RL, an experience replay buffer $\mathcal{B}$ is employed, which stores the agent's experiences $e_t=(s_t,a_t,r_t,s_{t+1})$ at each time step $t$. 
Meanwhile, a mini-batch of samples $(s,a,r,s^{\prime}) \sim U(\mathcal{B})$ will be drawn uniformly at random
from $\mathcal{B}$, and the following loss function will be calculated:
\begin{align}\label{eq.DQN}
L(\theta) = \mathbb{E}_{(s,a,r,s^{\prime}) \sim U(\mathcal{B})}\left[\left( r + \gamma \max\limits_{a\in\mathcal{A}} Q(s^\prime,a | \theta^{\prime}) - Q(s,a| \theta) \right)^2\right],
\end{align}
which can be used to update the network parameter by 
$\theta \leftarrow \theta - \alpha\cdot \nabla_{\theta}L(\theta)$ with a learning rate $\alpha$.
Note that in order to further improve the stability of RL, the original DNN in DQN utilizes a target network $\theta^{\prime}$ to derive the TD error for the agent, which adopts a so-called soft update strategy that tracks the weights of the learned network by $\theta^\prime \leftarrow \tau \theta + (1-\tau)\theta^\prime$ with $\tau \ll 1$.
Besides, the action taken by the agent at each time step $t$ is obtained by $a_t = \arg \max\limits_{a\in\mathcal{A}} Q(s_t,a | \theta)$.



\begin{figure}
\centering
\includegraphics[height = 8cm]{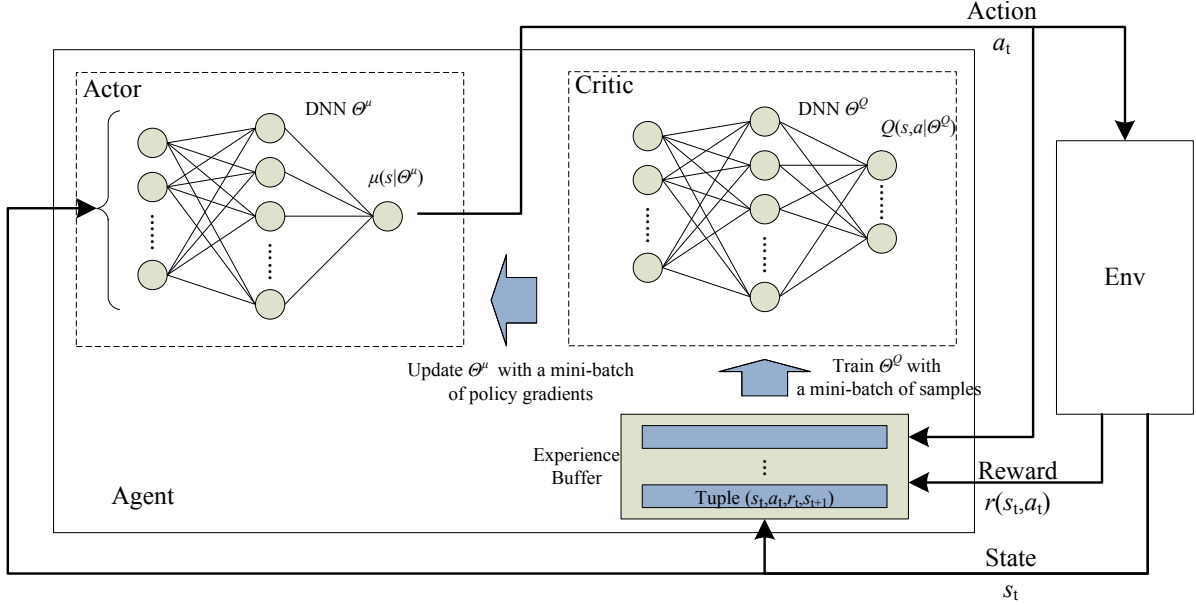}
\caption{Diagram of DDPG.}
\label{fig.ddpg}
\end{figure}

\subsection{DDPG}

Although problems in high-dimensional state spaces has been successfully solved by DQN, only discrete and low-dimensional action spaces can be handled. 
To extend DRL algorithms to continuous action space, DDPG has been proposed in~\cite{lillicrap2016continuous}. As shown in Fig.~\ref{fig.ddpg}, an actor-critic approach is adopted by using two separate DNNs to approximate the Q-value network $Q(s,a|\theta^Q)$, i.e., the critic function, and the policy network $\mu(s|\theta^{\mu})$, i.e., the actor function, respectively.~\footnote{Note that similar to DQN, both the original DNNs in DDPG also have its own target network parameterized by $\theta^{Q^\prime}$ and $\theta^{\mu^\prime}$, respectively, which use soft update strategy and slowly track the weights of the learned networks in the same way.} 
Specifically, the critic $Q(s,a|\theta^{Q})$ is similar to DQN and can be updated following \eqref{eq.DQN}.
On the other hand, the actor $\mu(s|\theta^{\mu})$ deterministically maps state $s$ to a specific continuous action. As derived in~\cite{silver2014deterministic}, the policy gradient of the actor can be calculated by chain rule,
\begin{align}\label{eq.PG}
\nabla_{\theta^{\mu}} J \approx \mathbb{E}_{(s,a,r,s^{\prime}) \sim U(\mathcal{B})}\left[\nabla_{a} Q(s,a|\theta^{Q}) \nabla_{\theta^{\mu}} \mu(s|\theta^\mu)\right],
\end{align}
which is the gradient of the expected return from the start distribution $J$ with respect to the actor parameter $\Theta^\mu$, 
averaged over the sampled mini-batch $U(\mathcal{B})$.
Thus, with \eqref{eq.DQN} and \eqref{eq.PG} in hand, network parameters of the actor and the critic can be updated by $\theta^{Q} \leftarrow \theta^{Q} - \alpha_Q\cdot \nabla_{\theta^Q}L(\theta^{Q})$ and $\theta^{\mu} \leftarrow \theta^{\mu} - \alpha_\mu\cdot \nabla_{\theta^\mu}J$, respectively. Here, $\alpha_Q$ and $\alpha_\mu$ are the learning rates. 

In order to improve the model with adequate exploration of the state space, one major challenge in RL is the tradeoff between exploration and exploitation~\cite{arulkumaran2017deep}, which is even more difficult for learning in continuous action spaces. 
As an off-policy algorithm, the exploration of DDPG can be treated independently from the learning process.
Thus, the exploration policy $\mu^\prime$ can be constructed by adding noise $\Delta\mu$ sampled ∫from a random noise process to the actor, i.e., 
\begin{align}
\mu^\prime(s) = \mu(s|\theta^{\mu}) + \Delta\mu,
\end{align}
where the random noise process needs to be elaborately selected. E.g., exploration noise sampled from a temporally correlated random process can better preserve momentum~\cite{lillicrap2016continuous}.





\section{Dynamic Computation Offloading for Mobile Edge Computing}\label{sec.MEC_Model}

\begin{figure}
\centering
\includegraphics[height = 6cm]{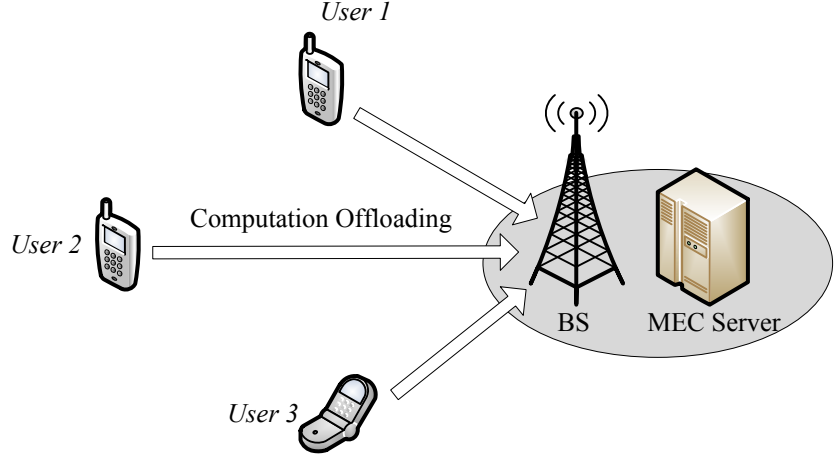}
\caption{Computation offloading in a multi-user mobile edge computing system.}
\label{fig.mec_model}
\end{figure}

As shown in Fig.~\ref{fig.mec_model}, we consider a multi-user MEC system, which consists of a BS, an MEC server and a set of mobile users $\mathcal{M} = \{1, 2, \ldots, M\}$. 
For each user $m \in \mathcal{M}$, it has computation-intensive tasks to be completed. 
Considering the limited computational resources on each mobile device, the MEC server is deployed in proximity to the BS by the telecom operator, which can improve user's computation experience by enabling it to offload part of its computation need to the MEC server via the wireless link~\cite{mach19mobile}.
A discrete-time model is adopted for the MEC system, where the operating period is slotted with equal length $\tau_0$ and indexed by $\mathcal{T} = \{0,1, \ldots\}$. The channel condition and task arrival of each user varies for each slot $t \in \mathcal{T}$. Thus, aiming to balance the average energy consumption and task processing delay, each user needs to determine the ratio of local execution and computation offloading at each slot. 
Moreover, as the number of mobile users increases, decentralized task scheduling at each user is more favorable, which can reduce the system overhead between the users and the MEC server and improve the scalability of the MEC system. 
In the following parts, we will introduce the modeling of networking and computing in detail. 


{
\begin{table}[t]
\center \protect
\caption{Summary of Notations}
\begin{tabular}{ll}
\hline
{\textbf{Notation}} & {\textbf{Description}}  \tabularnewline
\hline
{$\mathcal{M}$} & The set of mobile users   \tabularnewline
{$\mathcal{T}$} &Index set of the time slots \tabularnewline
{$\boldsymbol{h}_m(t)$} &Channel vector between user $m$ and the BS at slot $t$\tabularnewline
{$\boldsymbol{y}(t)$} & Received signal of the BS at slot $t$   \tabularnewline
{$\rho_m$} & Normalized temporal channel correlation coefficient of user $m$   \tabularnewline
{$\boldsymbol{H}(t)$} & Channel matrix from all the users to the BS at slot $t$   \tabularnewline
{$\boldsymbol{g}_m(t)$} & ZF detection vector for user $m$ at slot $t$   \tabularnewline
{$\gamma_m(t)$} & The receiving SINR of user $m$ at slot $t$   \tabularnewline
{$\lambda_m$} & Task arrival rate of user $m$  \tabularnewline
{$B_m(t)$} & Queue length of user $m$'s task buffer at slot $t$   \tabularnewline
{$a_m(t)$} & Number of task arrivals of user $m$ at slot $t$   \tabularnewline
{$f_m(t)$} & CPU frequency scheduled for local execution of user $m$ at slot $t$  \tabularnewline
{$L_m$ $(F_m)$} & CPU cycles required per one task bit (allowable CPU-cycle frequency) at user $m$   \tabularnewline

{$p_{o,m}(t)$} & Transmission power of user $m$ for computation offloading at slot $t$   \tabularnewline
{$d_{o,m}(t)$} & Data transmitted by user $m$ for computation offloading at slot $t$   \tabularnewline
{$p_{l,m}(t)$} & Power consumption of user $m$ for local execution at slot $t$   \tabularnewline
{$d_{l,m}(t)$} & Data processed by user $m$ via local execution at slot $t$   \tabularnewline
{$P_{o,m}$ $(P_{l,m})$} & Maximum transmission power (local execution power) of user $m$    \tabularnewline
{$\tau_{0}$} & Length of one time slot    \tabularnewline

\hline
\end{tabular}
\label{notationtable}
\end{table}
}
\subsection{Network Model}


In the MEC system, we consider a 5G macro-cell or small-cell BS, which is equipped with $N$ antennas and manages the uplink transmissions of multiple single-antenna mobile users by employing the well-known linear detection algorithm zero-forcing (ZF), which is of low complexity but efficient, especially for multi-user MIMO with large antenna arrays~\cite{ngo2013energy}. 
For each time slot $t \in \mathcal{T}$, if the channel vector of each mobile user $m \in \mathcal{M}$ is represented by $\boldsymbol{h}_{m}(t) \in \mathbb{C}^{N \times 1}$, the received signal of the BS can be written as
\begin{align}\label{eq.bs_receive}
\boldsymbol{y}(t) = \sum_{m=1}^{M} \boldsymbol{h}_m(t)\sqrt{p_{o,m}(t)} s_m(t) + \boldsymbol{n}(t),
\end{align}
where $p_{o,m}(t) \in [0, P_{o,m}]$ is the transmission power of user $m$ to offload task data bits with $P_{o,m}$ being the maximum value, $s_m(t)$ is the complex data symbol with unit variance, and $\boldsymbol{n}(t) \sim \mathcal{CN}(\boldsymbol{0}, \sigma_R^2\boldsymbol{I}_{N})$ is a vector of additive white Gaussian noise (AWGN) with variance $\sigma_R^2$.
Note that $\boldsymbol{I}_{N}$ denotes an $N \times N$ identity matrix. 
In order to characterize the temporal correlation between time slots for each mobile user $m \in \mathcal{M}$, the following Gaussian Markov block fading autoregressive model~\cite{suraweera2011effect} is adopted:
\begin{align}\label{eq.time_correlated_channel}
\boldsymbol{h}_m(t) = \rho_m \boldsymbol{h}_m(t-1) + \sqrt{1-\rho_m^2} \boldsymbol{e}(t),
\end{align}
where $\rho_m$ is the normalized channel correlation coefficient between slots $t$ and $t-1$, and the error vector $\boldsymbol{e}(t)$ is complex Gaussian and uncorrelated with $\boldsymbol{h}_m(t)$.
Note that $\rho_m = J_0(2 \pi f_{d,m} \tau_0)$ according to Jake's fading spectrum, where $f_{d,m}$ is the Doppler frequency of user $m$, $t_{0}$ is the slot length, and $J_0(\cdot)$ is the Bessel function of the first kind~\cite{abramowitz1972handbook}. 



Denoting $\boldsymbol{H}(t) = [\boldsymbol{h}_1(t),\ldots,\boldsymbol{h}_M(t)]$ as the $N \times M$ channel matrix between the BS and the $M$ users, the linear ZF detector at the BS~\footnote{Here, we consider the number of antennas at the BS is larger than the number of users, i.e., $N > M$. According to~\cite{ngo2013energy}, it is demonstrated that as $N$ increases, the SINR performance of ZF detection approaches that of the MMSE detection.} can be written by the channel matrix's pseudo inverse $\boldsymbol{H}^\dagger(t) = \left(\boldsymbol{H}^H(t) \boldsymbol{H}(t)\right)^{-1}\boldsymbol{H}^H(t)$. 
If the $m$-th row of $\boldsymbol{H}^\dagger(t)$ is represented by $\boldsymbol{g}^H_m(t)$, the received signal for user $m$ is $\boldsymbol{g}^H_m(t)\boldsymbol{y}(t) = \sqrt{p_{o,m}(t)}s_m(t) + \boldsymbol{g}^H_m(t)\boldsymbol{n}(t)$, since we have $\boldsymbol{g}^H_i(t)\boldsymbol{h}_j(t) = \delta_{ij}$ for ZF detection~\cite{ngo2013energy}. Here, $\delta_{ij} = 1$ when $i=j$ and $0$ otherwise. 
Thus, the corresponding signal-to-interference-plus-noise (SINR) can be derived by
\begin{align}\label{eq.sinr}
\gamma_m(t) = \frac{p_{o,m}(t)}{\sigma^2_R \|\boldsymbol{g}_m(t)\|^2} = \frac{p_{o,m}(t)}{\sigma^2_R \left[\left(\boldsymbol{H}^H(t) \boldsymbol{H}(t)\right)^{-1}\right]_{mm}},
\end{align}
where $[\boldsymbol{A}]_{mn}$ is the $(m,n)$-th element of matrix $\boldsymbol{A}$.
From~\eqref{eq.sinr}, it can be verified that each user's SINR will become worse as the number of users $M$ increases, which requires each user to allocate more power for task offloading. In the sequel, we will show how the user learns to adapt to the environment from the SINR feedbacks.


\subsection{Computation Model}


In this part, how each mobile user $m \in \mathcal{M}$ takes advantage of local execution or computation offloading to satisfy its running applications will be discussed. 
Without loss of generality, we use $a_m(t)$ to quantify the number of task arrivals during slot $t \in \mathcal{T}$, which can be processed starting from slot $t+1$ and is independent and identically distributed (i.i.d) over different time slots with mean rate $\lambda_m = \mathbb{E}[a_m(t)]$. 
Besides, we assume that the applications are fine-grained~\cite{kwak2015dream}. That is, some bits of the computation tasks denoted by $d_{l,m}(t)$ will be processed on the mobile device, and some other bits denoted by $d_{o,m}(t)$ will be offloaded to and executed by the MEC server. 
Thus, if $B_m(t)$ stands for the queue length of user $m$'s task buffer at the beginning of slot $t$, it will evolve as follows:
\begin{align}\label{eq.buffer_update}
B_m(t+1) = \left[B_m(t) - \left(d_{l,m}(t) + d_{o,m}(t)\right)\right]^+ + a_m(t), \forall t \in \mathcal{T},
\end{align}
where $B_m(0) = 0$ and $[x]^+ = \max(x, 0)$.

\subsubsection{Local computing}
In this part, we will show the amount of data bits being processed locally given the allocated local execution power $p_{l,m}(t) \in [0, P_{l,m}]$.
To start with, we assume that the number of CPU cycles required to process one task bit at user $m$ is denoted by $L_m$, which can be estimated through off-line measurement~\cite{miettinen2010energy}. 
By chip voltage adjustment using DVFS techniques~\cite{burd1996processor}, the CPU frequency scheduled for slot $t$ can be written by 
\begin{align}
f_m(t) = \sqrt[3]{p_{l,m}(t)/\kappa},
\end{align}
where $\kappa$ is the effective switched capacitance depending on the chip architecture. Note that $0 \leq f_m(t) \leq F_{m}$ with $F_{m} =\sqrt[3]{P_{l,m}(t)/\kappa}$ being the maximum allowable CPU-cycle frequency of user $m$'s device.
As a result, the local processed bits at $t$-th slot can be derived by:
\begin{align}\label{eq.local_D}
d_{l,m}(t) = \tau_0 f_m(t) L^{-1}_m.
\end{align}

\subsubsection{Edge computing}
To take advantage of edge computing, it is worth noting that the MEC server is usually equipped with sufficient computational resources, e.g., a high-frequency multi-core CPU. Thus, it can be assumed that different applications can be handled in parallel with a negligible processing latency, and the feedback delay is ignored due to the small sized computation output.
In this way, all the task data bits offloaded to the MEC server via the BS will be processed. 
Therefore, according to \eqref{eq.sinr} and given the uplink transmission power $p_{o,m}(t)$, the amount of offloaded data bits of user $m$ during slot $t$ can be derived by
\begin{align}
d_{o,m}(t) = \tau_0 W \log_2\left(1+\gamma_m(t)\right),
\end{align}
where $W$ is the system bandwidth and $\gamma_m(t)$ is obtained from \eqref{eq.sinr}.


\section{DRL Based Decentralized Dynamic Computation Offloading}\label{sec.DRL_design}

In this section, we will develop a DRL based approach to minimize the computation cost of each mobile user in terms of energy consumption and buffering delay in the proposed multi-user MEC system.
Specifically, by employing the DDPG algorithm, a decentralized dynamic computation offloading policy will be learned independently at each user, which selects an action, i.e., allocated powers for both local execution and computation offloading, upon the observation of the environment from its own perspective.
It is worth noting that each user has no prior knowledge of the MEC system, which means the number of users $M$, and statistics of task arrivals and wireless channels are unknown to each user agent and thus the online learning process is totally model-free. 
In the following, by adopting DDPG, the DRL framework for decentralized dynamic computation offloading will be introduced, where the state space, action space and reward function will be defined. Then, how to take advantage of the framework to train and test the decentralized policies is also presented. 


\begin{figure}
\centering
\includegraphics[height = 4cm]{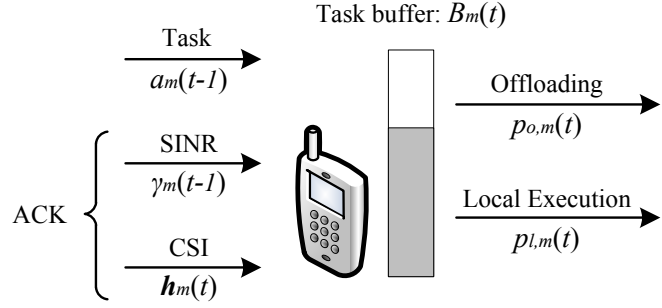}
\caption{Interaction of each user agent with the MEC system at slot $t$, including its local observation and selected action.}
\label{fig.agent}
\end{figure}

\subsection{The DRL Framework}

\textbf{\textit{State Space:}}
Full observation of the system includes the channel vectors and the queue lengths of the task buffer for all users. However, the system overhead to collect such informations at the BS and then distribute them to each user is very high in practice, which will become even higher as the number of mobile users increases. 
In order to reduce the overhead and make the MEC system much more scalable, we assume that the state of each user agent is only determined by its local observation of the system, upon which each user will select an action independently from other user agents. 

As shown in Fig.~\ref{fig.agent}, at the start of time slot $t$, the queue length of each user $m$'s data buffer $B_{m}(t)$ will be updated according to \eqref{eq.buffer_update}. Meanwhile, one feedback from the BS conveying the last receiving SINR of user $m$ at the BS, i.e., $\gamma_m(t-1)$, will be received.
At the same time, channel vector $\boldsymbol{h}_m(t)$ for the upcoming uplink transmission can be estimated by using channel reciprocity.
As a result, from the perspective of each user $m$, the state can be defined as 
\begin{align}
s_{m,t} = \left[B_{m}(t), \phi_m(t-1), \boldsymbol{h}_m(t)\right],
\end{align}
where we denote the projected power ratio after ZF detection at the BS for slot $t$ as 
\begin{align}
\phi_m(t) = \frac{\gamma_m(t) \sigma_R^2}{p_{o,m}(t)\|\boldsymbol{h}_m(t)\|^2} = \frac{1}{\|\boldsymbol{h}_m(t)\|^2 \left[\left(\boldsymbol{H}^H(t) \boldsymbol{H}(t)\right)^{-1}\right]_{mm}}.
\end{align}
Note that in order to decode user $m$'s symbol without inter-stream interference, ZF detection will project the received signal $\boldsymbol{y}(t)$ to the subspace orthogonal to the one spanned by the other users' channel vectors~\cite{tse2005fundamentals}. 
In this way, $\phi_m(t)$ can be interpreted as the ratio of unit received power of user $m$'s uplink signal after projection.



\textbf{\textit{Action Space:}}
Based on the current state $s_{m,t}$ of the system observed by each user agent $m$, an action $a_{m,t}$ including the allocated powers for both local execution and computation offloading will be selected for each slot $t \in \mathcal{T}$ as below:
\begin{align}
a_{m,t} = \left[p_{l,m}(t),p_{o,m}(t)\right].
\end{align}
It is worth noting that, by applying the DDPG algorithm, either power allocation can be elaborately optimized in a continuous action space, i.e., $p_{l,m}(t) \in [0, P_{l,m}]$ and $p_{o,m}(t) \in [0, P_{o,m}]$, to minimize the average computation cost, unlike other conventional DRL algorithms to select from several predefined discrete power levels. 
Consequently, the high dimension of discrete action spaces can be significantly reduced. 

\textbf{\textit{Reward Function:}}
As mentioned in Section \ref{sec.pre}, the behavior of each user agent is reward-driven, which indicates that the reward function plays a key role in the performance of DRL algorithms. 
In order to learn an energy-aware dynamic computation offloading policy for the proposed MEC model, we consider to minimize the
energy consumption while completing tasks within a acceptable buffering delay. 
Thus, the overall computation cost for each user agent will be counted by both the total energy cost and the penalty on task buffering delay. 
Notice that it can be known that according to the Little's Theorem~\cite{shortle2018fundamentals}, the average queue length of the task buffer is proportional to the buffering delay. 
In this way, we define the reward function $r_{m,t}$ that each user agent $m$ receives after slot $t$ as 
\begin{align}\label{eq.reward_original}
r_{m,t} = -w_{m,1} \cdot \left(p_{l,m}(t) + p_{o,m}(t)\right) - w_{m,2} \cdot B_{m}(t),
\end{align}
where $w_{m,1}$ and $w_{m,2}$ are both nonnegative weighted factors, and the reward $r_{m,t}$ is the negative wighted sum of the instantaneous total power consumption and the queue length of task buffer.
By setting different values of $w_{m,1}$ and $w_{m,2}$, a tradeoff between energy consumption and buffering delay can be made for the dynamic computation offloading policy. 
Moreover, it is worth noting that although the DDPG based algorithm maximizes value function of user $m$ starting from the initial state $s_{m,1}$ under policy $\mu_m$, i.e., 
\begin{align}
V^{\mu_m}(s_{m,1}) = \mathbb{E} \left[\sum_{t=1}^\infty \gamma^{t-1} r_{m,t} | s_{m,1} \right], 
\end{align}
it can be used to approximate of the real expected infinite-horizon undiscounted return~\cite{adelman2008relaxations} at each user agent when $\gamma \rightarrow 1$. That is, the following average computation cost 
\begin{align}
\bar{C}_m(s_{m,t})= \mathbb{E} \left[\lim_{T \rightarrow \infty}\frac{1}{T}\sum_{i=t}^T  w_{m,1} \left(p_{l,m}(i) + p_{o,m}(i)\right) + w_{m,2} B_{m}(i) | s_{m,t} \right],
\end{align}
will be minimized by applying the learned computation offloading policy $\mu^*_m$.





\subsection{Training and Testing}
To learn and evaluate the decentralized computation offloading algorithm, there are two stages including training and testing for the DRL framework. 
Before training, we need to build the DRL framework with a simulated environment and a group of user agents. 
The simulated environment is used to generate the training and testing data, which mimics the interaction of the user agents with the MEC system needs to be built, which accepts the decision of each user agent and returns feedbacks of CSI and SINR. 
For each user agent, the TensorFlow library~\cite{tensorflow2015-whitepaper} is used to construct and train the DNNs in the DDPG algorithm. 

The detailed training stage is illustrated in Algorithm~\ref{ag.1}. 
It is worth noting that the interaction between the user agents and the environment is generally continuing RL tasks~\cite{sutton1998reinforcement}, which does not break naturally into identifiable episodes.~\footnote{Note that for episodic RL tasks, each initial state of the user agent will terminate at a specific state.}
Thus, in order to have better exploration performance, the interaction of the user agent in the MEC system will be manually start with a random initial state $s_{m,1}$ and terminate at a predefined maximum steps $T_{\max}$ for each episode. 
At each time step $t$ during an episode, each agent's experience tuple $(s_{m,t}, a_{a,t}, r_{m,t}, s_{m,t+1})$ will be stored in its own experience buffer $\mathcal{B}_m$. Meanwhile, the use agent's actor and critic network will be updated accordingly using a mini-batch of experience tuples $\{(s_i,a_i,r_i,s^\prime_i)\}_{i=1}^I$ randomly sampled from the replay buffer $\mathcal{B}_m$. 
In this way, after the training of $K_{\max}$ episodes, the dynamic computation offloading policy will be gradually and independently learned at each user agent. 

As for the testing stage, each user agent will firstly load its actor network parameters learned in the training stage. Then, the user agent will start with an empty data buffer and interact with a randomly initialized environment, after which it selects actions according to the output of the actor network, when its local observation of the environment is obtained as the current state. 

\begin{algorithm} 
\caption{Training Stage for the DDPG based Dynamic Computational Offloading}\label{ag.1}
\begin{algorithmic}[1]
\For{each user agent $m \in \mathcal{M}$}
\State Randomly initialize the actor network $\mu(s|\theta_m^\mu)$ and the critic network $Q(s,a|\theta_m^{Q})$;
\State Initialize the associated target networks with weights $\theta_m^{\mu^\prime} \leftarrow \theta_m^{\mu} and$ $\theta_m^{Q^\prime} \leftarrow \theta_m^{Q}$;
\State Initialize the experience replay buffer $\mathcal{B}_m$; 
\EndFor
\For{each episode $k = 1,2,\ldots, K_{\max}$}
\State Reset simulation parameters for the multi-user MEC model environment;
\State Randomly generate an initial state $s_{m,1}$ for each user agent $m \in \mathcal{M}$;
\For{each time slot $t = 1,2,\ldots,T_{\max}$}
\For{each user agent $m \in \mathcal{M}$}
\State Determine the power for local execution and computation offloading by selecting an action $a_{m,t} = \mu(s_{m,t}|\theta_m^\mu) + \Delta\mu$ using running the current policy network $\theta_m^\mu$ and generating exploration noise $\Delta\mu$;
\State Execute action $a_{m,t}$ independently at the user agent, and then receive reward $r_{m,t}$ and observe the next state $s_{m,t+1}$ from the environment simulator;
\State Collect and save the tuple $(s_{m,t},a_{m,t},r_{m,t},s_{m,t+1})$ into the replay buffer $\mathcal{B}_m$;
\State Randomly sample a mini-batch of $I$ tuples $\{(s_i,a_i,r_i,s^\prime_i)\}_{i=1}^I$ from $\mathcal{B}_m$;
\State Update the critic network $Q(s,a|\theta_m^{Q})$ by minimizing the loss $L$ with the samples:
\begin{align}
L = \frac{1}{I}\sum_{i=1}^I \left( r_i + \max_{a\in\mathcal{A}} Q(s_i^\prime, a | \theta_m^{Q^\prime}) - Q(s_i,a_i | \theta_m^Q) \right)^2;  
\end{align}
\State Update the actor network $\mu(s|\theta_m^\mu)$ by using the sampled policy gradient:
\begin{align}
\nabla_{\theta_m^{\mu}} J \approx \frac{1}{I} \sum_{i=1}^I \nabla_{a} Q(s_i,a|\theta_m^{Q}) |_{a=a_i} \nabla_{\theta_m^{\mu}} \mu(s_i|\theta_m^\mu);
\end{align}
\State Update the target networks by $\theta_m^{\mu^\prime} \leftarrow \tau \theta_m^{\mu} + (1-\tau)\theta_m^{\mu^\prime}$ and $\theta_m^{Q^\prime} \leftarrow \tau \theta_m^{Q} + (1-\tau)\theta_m^{Q^\prime}$;
\EndFor
\EndFor
\EndFor
\end{algorithmic}
\end{algorithm}

\section{Numerical Results}\label{sec.simulation}




In this section, numerical simulations will be presented to illustrate the proposed DRL framework for decentralized dynamic computation offloading in the MEC system. System setup for the simulations is firstly introduced. Then, performance of the DRL framework is demonstrated and compared with some other baseline schemes in the scenarios of single user and multiple users, respectively. 

\subsection{Simulation Setup}

In the MEC system, time is slotted by $\tau_0 = 1 \mathrm{ms}$. 
For the beginning of every episode, each user $m$'s channel vector is initialized as $\boldsymbol{h}_m(0) \sim \mathcal{CN}(0, h_0(d_0/d_m)^{\alpha}\boldsymbol{I}_N)$, where the path-loss constant $h_0 = -30\mathrm{dB}$, the reference distance $d_0 = 1\mathrm{m}$, the path-loss exponent $\alpha = 3$, and $d_m$ is the distance of user $m$ to the BS in meters.
In the following slots, $\boldsymbol{h}_m(t)$ will be updated according to \eqref{eq.time_correlated_channel}, where the channel correlation coefficient $\rho_m = 0.95$ and the error vector $\boldsymbol{e}(t) \sim \mathcal{CN}(0, h_0(d_0/d)^{\alpha}\boldsymbol{I}_N)$ with $f_{d,m} = 70\mathrm{Hz}$.
Additionally, we set the system bandwidth to be $1\mathrm{MHz}$, the maximum transmission power $P_{o,m} = 2\mathrm{W}$, and the noise power $\sigma_R^2 = 10^{-9} \mathrm{W}$. 
On the other hand, for local execution, we assume that $\kappa = 10^{-27}$, the required CPU cycles per bit $L_m = 500$ cycles/bit, and the maximum allowable CPU-cycle frequency $F_m = 1.26\mathrm{GHz}$, from which we know that the maximum power required for local execution $P_{l,m} = 2\mathrm{W}$.

To implement the DDPG algorithm, for each user agent $m$, the actor network and critic network is a four-layer fully connected neural network with two hidden layers. The number of neurons in the two hidden layers are $400$ and $300$, respectively. The neural networks use the Relu, i.e., $f(x) = \max(0, x)$, as the activation function for all hidden layers, while the final output layer of the actor uses a sigmoid layer to bound the actions. Note that for the critic, actions are not included until the second hidden layer of the Q-network. 
Adaptive moment estimation (Adam) method~\cite{kingma2014adam} is used for learning the neural network parameters with a learning rate of $0.0001$ and $0.001$ for the actor and critic respectively. 
The soft update rate for the target networks is $\tau = 0.001$.
To initialize the network layer weights, settings in experiment of \cite{lillicrap2016continuous} is adopted. 
Moreover, in order to explore well, the Ornstein-Uhlenbeck process~\cite{uhlenbeck1930theory} with $\theta=0.15$ and $\sigma=0.12$ is used to provide temporal correlated noise. The experience replay buffer size is set as $|\mathcal{B}_m| = 2.5 \times 10^5$. 

With slight abuse of notation, we introduce a tradeoff factor $w_m \in [0,1]$ for each use agent $m \in \mathcal{M}$, which is used to represent the two nonnegative weighted factors by $w_{m,1} = 10w_{m}$ and $w_{m,2} = 1-w_m$. Thus, the reward function $r_{m,t}$ in \eqref{eq.reward_original} can be written by
\begin{align}\label{eq.reward_simple}
r_{m,t} = -10w_{m}\cdot \left(p_{l,m}(t) + p_{o,m}(t)\right) - (1-w_{m}) \cdot B_{m}(t),
\end{align}
from which we can make a tradeoff between energy consumption and buffering delay by simply setting a single factor $w_m$. 
Moreover, the number of episodes is $K_{\max} = 2000$ in the training stage, and the maximum steps of each episode is $T_{\max} = 200$. 

For comparison, the baseline strategies are introduced as follows:
\subsubsection{Greedy Local Execution First (GD-Local)}
For each slot, the user agent firstly attempts to execute task data bits locally as many as possible. Then, the remaining buffered data bits will be offloaded to the MEC. 

\subsubsection{Greedy Computation Offloading First (GD-Offload)}
Similar to the GD-local strategy, each user agent firstly makes its best effort to offload data bits to the MEC, and then local execution will be adopted to process the remaining buffered data bits for each time slot.

\subsubsection{DQN based Dynamic Offloading (DQN)}
To evaluate performance of the proposed DDPG base algorithm, the conventional discrete action space based DRL algorithm, i.e., DQN~\cite{mnih2015human}, is also implemented for the dynamic computation offloading problem.
Specifically, for each user $m$, the power levels for local execution and computation offloading are defined as $\mathcal{P}_{l,m} = \{0, \frac{P_{l,m}}{L-1}, \dots, P_{l,m}\}$ and $\mathcal{P}_{o,m} = \{0, \frac{P_{o,m}}{L-1}, \dots, P_{o,m}\}$, where the number of power levels is set as $L = 8$. 
Thus, the action space for each user agent to select from is $\mathcal{P}_{l,m} \times \mathcal{P}_{o,m}$. Besides, $\epsilon$-greedy exploration and Adam method are adopted for training. 

\subsection{Single User Scenario}
In this part, numerical results of training and testing for the single user scenario are illustrated. The user is assumed to be randomly located in a distance of $d_1 = 100$ meters to the BS. 

\subsubsection{Training}
\begin{figure}[tbp]
\centering
\subfigure[$w_1 = 0.5$]{
\centering
\includegraphics[height = 6.5cm]{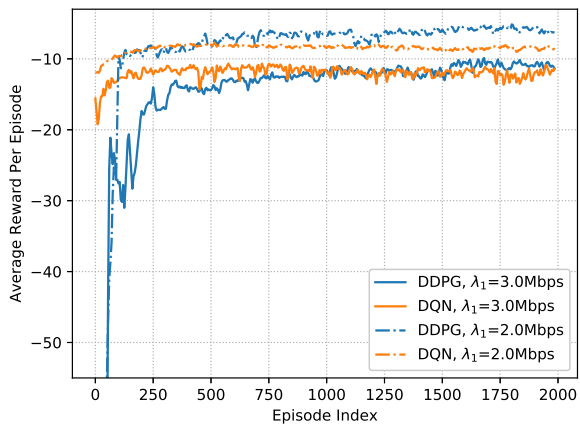}
\label{fig.t_05_nB_reward_train}}
\subfigure[$w_1 = 0.8$]{
\centering
\label{fig.t_08_nB_reward_train}
\includegraphics[height = 6.5cm]{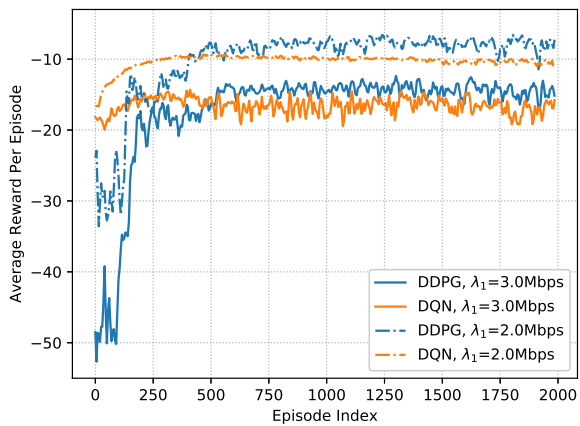}
}
\caption{Illustration of the average reward per episode in the training process for a single user agent.}
\end{figure}

As shown in Fig. \ref{fig.t_05_nB_reward_train} and Fig. \ref{fig.t_08_nB_reward_train}, the training process of the single-user dynamic computation offloading is presented by setting $w_1 = 0.5$ and $w_1 = 0.8$, respectively. 
Note that the results are averaged from $10$ runs of numerical simulations. 
In each figure, we will compare two different cases, where the task arrival rate is set as $\lambda_1 = 2.0 \mathrm{Mbps}$ and $\lambda_1 = 3.0 \mathrm{Mbps}$, respectively. 
It can be observed that for both policies learned from DDPG and DQN, the average reward of each episode increases as the interaction between the user agent and the MEC system environment continues, which indicates that efficient computation offloading policies can be successfully learned without any prior knowledge. 
Besides, the performance of each learned policy becomes stable after about $1500$ episodes. 
On the other hand, performance of the policy learned from DDPG is always better than DQN for different scenarios, which demonstrates that for continuous control problems, DDPG based strategies can explore the action space more efficiently than DQN based strategies. 

\label{fig.reward_single_train}

\subsubsection{Testing}

\begin{figure*}[tbp]
\centering
\includegraphics[height = 6cm]{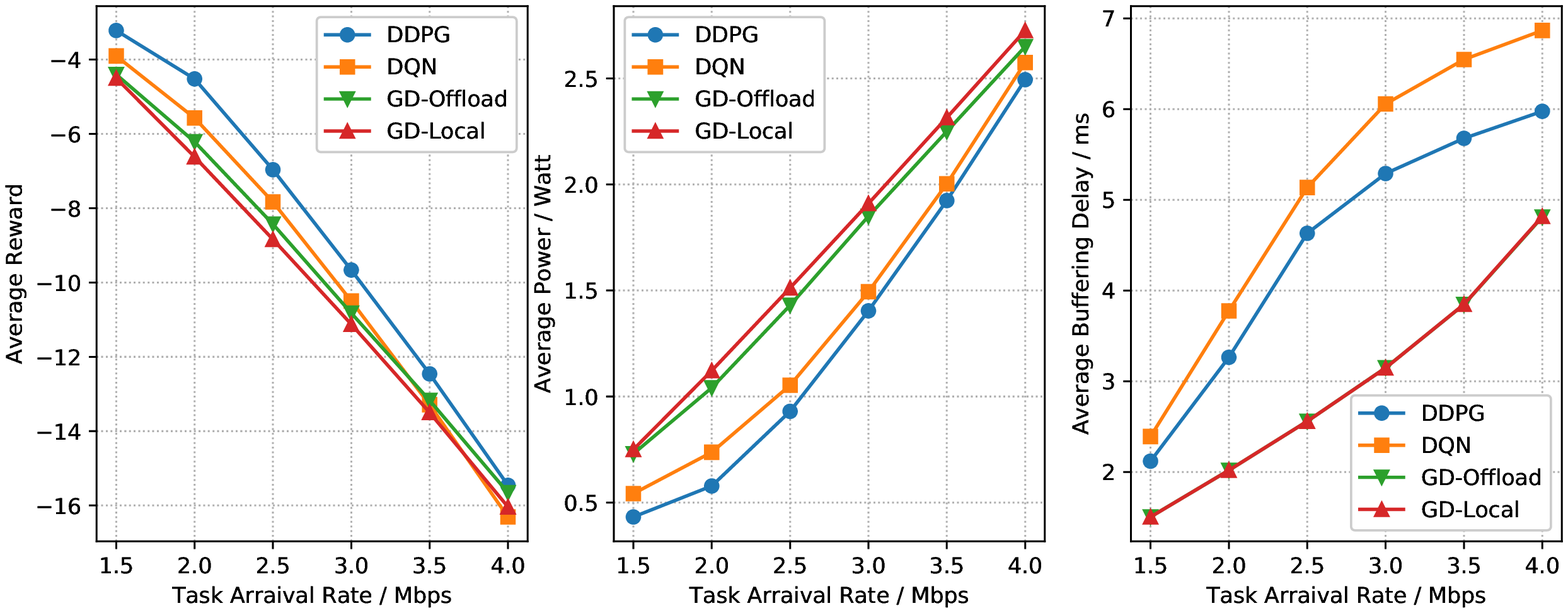}
\caption{Comparison of testing results for a single user agent with $w_1 = 0.5$.}
\label{fig.t_05_nB_test}
\end{figure*}

\begin{figure*}[tbp]
\centering
\includegraphics[height = 6cm]{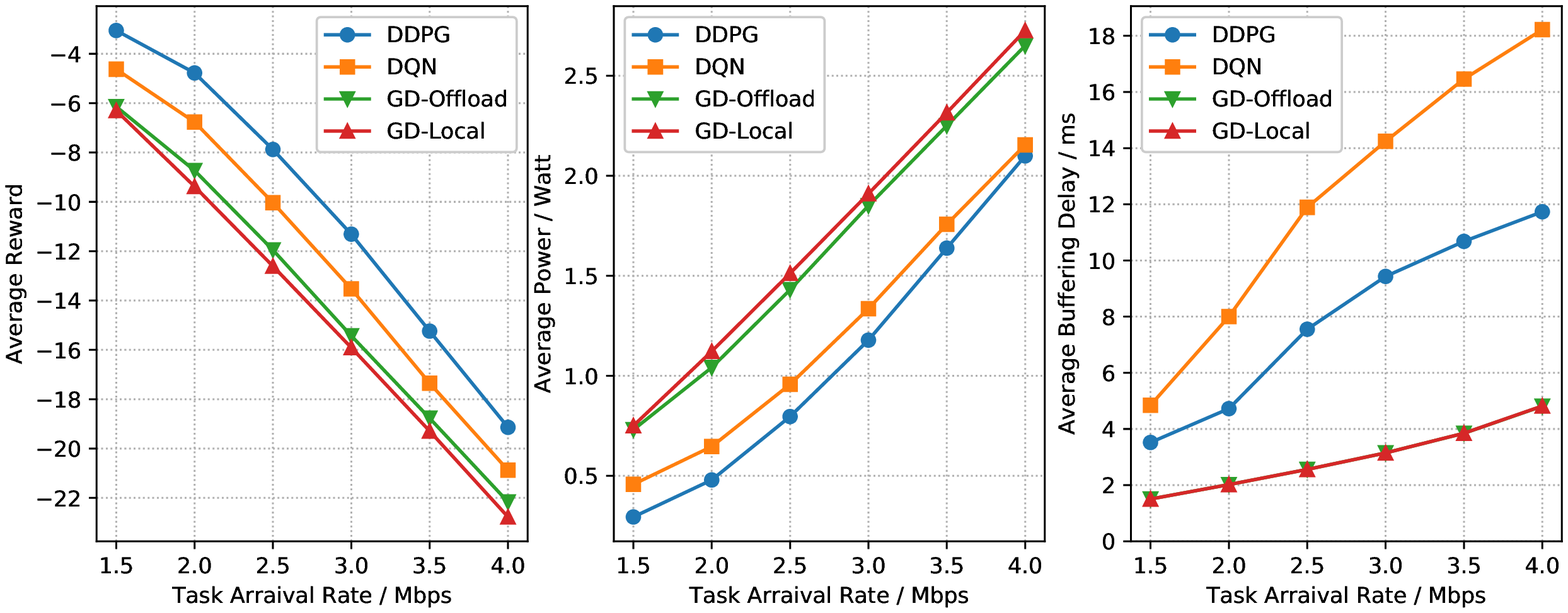}
\caption{Comparison of testing results for a single user agent with $w_1 = 0.8$.}
\label{fig.t_08_nB_test}
\end{figure*}

In the training stage, we have obtained dynamic computation offloading policies by applying the DDPG based and DQN based learning algorithms after $K_{\max} = 2000$ episodes, respectively.
For different task arrival rates ranging from $\lambda_1 = 1.5 \sim 4.0 \mathrm{Mbps}$, the actor and critic network will be trained with the same network architecture and hyper-parameters. 
To compare the performance of different policies, testing results are averaged from $100$ runs of numerical simulations, where each run consists of $10000$ steps. 
As shown in Fig. \ref{fig.t_05_nB_test} and Fig. \ref{fig.t_08_nB_test}, testing results are presented for $w_1 = 0.5$ and $w_1 = 0.8$, respectively, each of which includes the performance of average reward, power consumption and buffering delay.  
It can be observed from Fig. \ref{fig.t_05_nB_test} that the average reward will increase as the task arrival rate grows, which indicates that the computation cost is higher for a larger computation demand. 
Specifically, the increased computation cost results from a higher power consumption and a longer buffering delay. 
Moreover, although the DDPG based strategy outperforms both greedy strategies with minimum average reward, which, however, slightly compromises the buffering delay to achieve the lowest energy consumption.
It is worth noting that the average reward of the DQN based strategy is higher than the greedy strategies, which is due to the limited number of discrete power levels of DQN based strategy\footnote{Note that although finer grained discretization of the action space will potentially lead to better performance, the number of actions increases exponentially with the number of degrees of freedom, which makes it much more challenging to explore efficiently and in turn significantly decreases the performance.}.

In Fig. \ref{fig.t_08_nB_test}, testing results for a larger tradeoff factor $w_1 = 0.8$ are also provided. 
From \eqref{eq.reward_simple}, we know that a larger $w_1$ will give more penalty on power consumption in the reward function, i.e., the computation cost. 
In this scenario, the average reward of the DDPG based strategy  outperforms all the other strategies, and the gap is much larger than that in the case of $w_1 = 0.5$. 
Specifically, this is achieved by much lower power consumption and increased buffering delay. 



\begin{figure}[tbp]
\centering
\includegraphics[height = 6.5cm]{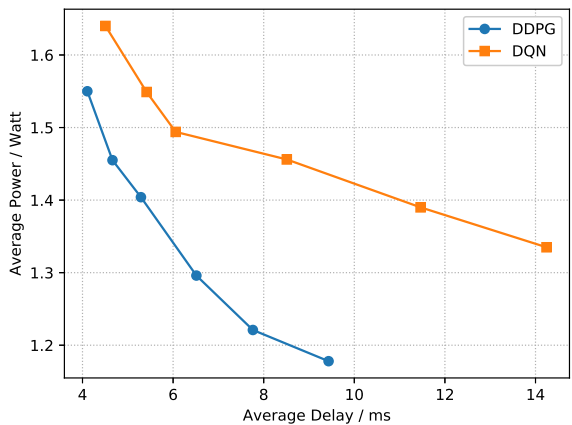}
\caption{Illustration of power-delay tradeoff for a single user agent with $\lambda_1 = 2.0\mathrm{Mbps}$. Note that for the points on each cuve from left to right, the tradeoff factor $w_1$ varies from $0.3 \sim 0.8$, correspondingly. }
\label{fig.tradeoff}
\end{figure}

\subsubsection{Power-Delay Tradeoff}
We also investigate testing results for the power-delay tradeoff by setting different values of $w_1$ in Fig. \ref{fig.tradeoff}.
It can be inferred from the curves that, there is a tradeoff between the average power consumption and the average buffering delay. Specifically, with a larger $w_1$, the power consumption will be decreased by sacrificing the delay performance, which indicates that in practice $w_1$ can be tuned to have a minimum power consumption with a given delay constraint. 
It is also worth noting noting that for each value of $w_1$, the policy learned form DDPG always has better performance in terms of both power consumption and buffering delay, which demonstrates the superiority of the DDPG based strategy for continuous power control.

\subsection{Multi-User Scenario}

In this part, numerical results for the multi-user scenario is presented. There are $M=3$ mobile users in the MEC system, each of which is randomly located in a distance of $d_m = 100$ meters to the BS, and the task arrival rate is $\lambda_m = m \times 1.0 \mathrm{Mbps}$, for $m \in \{1,2,3\}$. 

\subsubsection{Training}
\begin{figure}[tbp]
\centering
\includegraphics[height = 6.5cm]{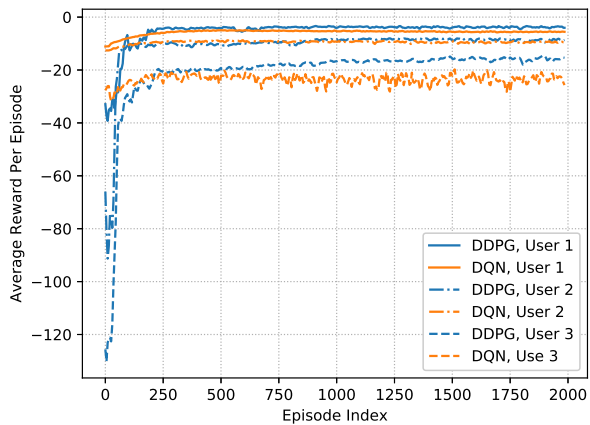}
\caption{Illustration of the average reward in the training process with $w_m = 0.5$ for all mobile users.}
\label{fig.t_M_05_nB_reward_train}
\end{figure}

By setting $w_m = 0.5$ for all the users, the training process has been shown in Fig. \ref{fig.t_M_05_nB_reward_train}. It can be observed that for each mobile user, the average reward increases gradually when the mobile user interacts with the MEC system after more episodes. 
Thus, we know that for both the DDPG based and DQN based strategies, efficient decentralized dynamic computation offloading policies can be learned at each mobile user, especially for heterogeneous users with different computation demands.
Moreover, it can be inferred that the higher computation cost needs to be paid by the user with a higher computation demand.
Meanwhile, compared with the single user scenario, the average reward obtained in the multi-user scenario is much lower for the same task arrival rate. 
It is due to the fact that the spectral efficiency of data transmission will be degraded when more mobile users are served by the BS. Hence, more power will be consumed in computation offloading in the multi-user scenario. 

\subsubsection{Testing}
By loading the neural network parameters learned by the DDPG based and DQN based algorithms after $K_{\max} = 2000$ episodes, testing results of different dynamic computation offloading policies are compared in Table \ref{t.t05} and Table \ref{t.t08}. 
From Table \ref{t.t05}, we know that the average rewards of user $2$ and user $3$ adopting DDPG based strategies are better than all other strategies under the scenario of $w_m = 0.5$. However, as for user $1$, the DDPG based strategy is slightly worse than the GD-Local strategy, which indicates that the exploration of DDPG for a small allocated power needs to be further improved. 
Also, it can be observed that both the DDPG based and DQN based strategies can achieve much lower power consumption with a little compromised buffering delay. 

By setting the tradeoff factor $w_m = 0.8$ as shown in Fig. \ref{t.t08}, the DDPG based strategies obtain the best average reward at each user agent. Moreover, the performance gaps between the DDPG based strategies and other greedy strategies become larger. 
Besides, with more penalty given to the power consumption, the consumed power of each user is much lower than that of $w_m = 0.5$, which, however, results in a moderately increased buffering delay. 
Thus, we know that for the multi-user scenario, power consumption can be minimized with a satisfied average buffering delay, by selecting a proper value of $w_m$. 
Notice that again the DDPG based strategies outperform the DQN based strategies in terms of average reward for all users. 

\begin{table}
\caption{Comparison of testing results with $w_m = 0.5$ for all mobile users.}
\label{t.t05}
\centering
\begin{tabular}{ c ||c  c c|| c  c c|| ccc}
  \hline
      & \multicolumn{3}{c||}{Average Reward}& \multicolumn{3}{c||}{Average Power}& \multicolumn{3}{c}{Average Delay} \\
      & \emph{User 1} & \emph{User 2} & \emph{User 3} & \emph{User 1} & \emph{User 2} & \emph{User 3}  & \emph{User 1} & \emph{User 2} & \emph{User 3} \\
  \hline
  DDPG & -1.770 & \textbf{-5.670} & \textbf{-12.782} & \textbf{0.205} & \textbf{0.774} & \textbf{1.939} & 1.489 & 3.600 & 6.174 \\
  DQN &  -2.174 & -7.657 & -14.688  & 0.292 & 1.156 & 2.320  & 1.428 & 3.753 & 6.176 \\
  GD-Offload &  -2.514 & -7.597 & -18.690 &  0.402 & 1.309 & 3.143 & \textbf{1.007} & \textbf{2.103} & \textbf{5.951}  \\
  GD-Local & \textbf{-1.633} & -9.504 & -20.071 &  0.216 & 1.678 & 3.407 &  1.106 & 2.228 & 6.072 \\
  \hline
\end{tabular}
\end{table}

\begin{table}
\caption{Comparison of testing results with $w_m = 0.8$ for all mobile users.}
\label{t.t08}
\centering
\begin{tabular}{ c ||c  c c|| c  c c|| ccc}
  \hline
      & \multicolumn{3}{c||}{Average Reward}& \multicolumn{3}{c||}{Average Power}& \multicolumn{3}{c}{Average Delay} \\
      & \emph{User 1} & \emph{User 2} & \emph{User 3} & \emph{User 1} & \emph{User 2} & \emph{User 3}  & \emph{User 1} & \emph{User 2} & \emph{User 3} \\
  \hline
  DDPG & \textbf{-1.919} & \textbf{-6.366} & \textbf{-16.164} &\textbf{0.162} & \textbf{0.602} & \textbf{1.674}  & 3.114 & 7.752 & 13.861\\
  DQN &  -2.780 & -8.915 & -18.675  & 0.284 & 0.915 & 1.954 &2.539 & 7.973 & 15.216 \\
  GD-Offload &  -3.417 & -10.893 & -26.334 &  0.402 & 1.309 & 3.143 & \textbf{1.007} & \textbf{2.103} & \textbf{5.951}  \\
  GD-Local &-1.949 & -13.870 & -28.470 &  0.216 & 1.678 & 3.407 &  1.106 & 2.228 & 6.072 \\
  \hline
\end{tabular}
\end{table}

\section{Conclusion}\label{sec.conculusion}
In this paper, we considered a multi-user MEC system, where tasks arrive stochastically and wireless channels are time-varying at each user. 
In order to minimize the long-term average computation cost in terms of power consumption and buffering delay, the design of DRL based decentralized dynamic computation offloading algorithms has been investigated. 
Specifically, by adopting the continuous action space based DRL approach named DDPG, an efficient computation offloading policy has been successfully learned at each mobile user, which is able to allocate powers of local execution and task offloading adaptively from its local observation of the MEC system.
Numerical simulations have been performed to verify the superiority of the proposed DDPG based decentralized strategy over the conventional DQN based discrete power control strategy and some other greedy strategies with reduced computation cost. 
Besides, the power-delay tradeoff for both the DDPG based and DQN based strategies has been also studied.


\begin{thebibliography}{10}
\providecommand{\url}[1]{#1}
\csname url@samestyle\endcsname
\providecommand{\newblock}{\relax}
\providecommand{\bibinfo}[2]{#2}
\providecommand{\BIBentrySTDinterwordspacing}{\spaceskip=0pt\relax}
\providecommand{\BIBentryALTinterwordstretchfactor}{4}
\providecommand{\BIBentryALTinterwordspacing}{\spaceskip=\fontdimen2\font plus
\BIBentryALTinterwordstretchfactor\fontdimen3\font minus
  \fontdimen4\font\relax}
\providecommand{\BIBforeignlanguage}[2]{{%
\expandafter\ifx\csname l@#1\endcsname\relax
\typeout{** WARNING: IEEEtran.bst: No hyphenation pattern has been}%
\typeout{** loaded for the language `#1'. Using the pattern for}%
\typeout{** the default language instead.}%
\else
\language=\csname l@#1\endcsname
\fi
#2}}
\providecommand{\BIBdecl}{\relax}
\BIBdecl

\bibitem{zhang2013toward}
W.~Zhang, Y.~Wen, J.~Wu, and H.~Li, ``Toward a unified elastic computing
  platform for smartphones with cloud support,'' \emph{IEEE Netw}, vol.~27,
  no.~5, pp. 34--40, 2013.

\bibitem{sun2016edgeiot}
X.~Sun and N.~Ansari, ``{EdgeIoT}: Mobile edge computing for the internet of
  things,'' \emph{IEEE Commun. Mag.}, vol.~54, no.~12, pp. 22--29, 2016.

\bibitem{zhang2017mobile}
K.~Zhang, Y.~Mao, S.~Leng, Y.~He, and Y.~Zhang, ``Mobile-edge computing for
  vehicular networks: A promising network paradigm with predictive
  off-loading,'' \emph{IEEE Veh. Technol. Mag}, vol.~12, no.~2, pp. 36--44,
  2017.

\bibitem{satyanarayanan2017emergence}
M.~Satyanarayanan, ``The emergence of edge computing,'' \emph{Computer},
  vol.~50, no.~1, pp. 30--39, 2017.

\bibitem{shi2016edge}
W.~Shi, J.~Cao, Q.~Zhang, Y.~Li, and L.~Xu, ``Edge computing: Vision and
  challenges,'' \emph{IEEE Internet of Things Journal}, vol.~3, no.~5, pp.
  637--646, 2016.

\bibitem{mach19mobile}
P.~Mach and Z.~Becvar, ``Mobile edge computing: A survey on architecture and
  computation offloading,'' \emph{IEEE Commun. Surveys Tuts.}, vol.~19, no.~3,
  pp. 1628--1656, 2017.

\bibitem{mao2017survey}
Y.~Mao, C.~You, J.~Zhang, K.~Huang, and K.~B. Letaief, ``A survey on mobile
  edge computing: The communication perspective,'' \emph{IEEE Commun. Surveys
  Tuts.}, vol.~19, no.~4, pp. 2322--2358, 2017.

\bibitem{patel2014mobile}
M.~Patel, B.~Naughton, C.~Chan, N.~Sprecher, S.~Abeta, A.~Neal \emph{et~al.},
  ``Mobile-edge computing introductory technical white paper,'' \emph{White
  Paper, Mobile-edge Computing (MEC) industry initiative}, 2014.

\bibitem{kumar2010cloud}
K.~Kumar and Y.-H. Lu, ``Cloud computing for mobile users: Can offloading
  computation save energy?'' \emph{Computer}, vol.~43, no.~4, pp. 51--56, 2010.

\bibitem{kosta2012thinkair}
S.~Kosta, A.~Aucinas, P.~Hui, R.~Mortier, and X.~Zhang, ``Thinkair: Dynamic
  resource allocation and parallel execution in the cloud for mobile code
  offloading,'' in \emph{Proc. IEEE INFOCOM}, 2012, pp. 945--953.

\bibitem{huang2012dynamic}
D.~Huang, P.~Wang, and D.~Niyato, ``A dynamic offloading algorithm for mobile
  computing,'' \emph{IEEE Trans. Wireless Commun.}, vol.~11, no.~6, pp.
  1991--1995, 2012.

\bibitem{chen2018task}
M.~Chen and Y.~Hao, ``Task offloading for mobile edge computing in software
  defined ultra-dense network,'' \emph{IEEE J. Sel. Areas Commun.}, vol.~36,
  no.~3, pp. 587--597, 2018.

\bibitem{chen2015decentralized}
X.~Chen, ``Decentralized computation offloading game for mobile cloud
  computing,'' \emph{IEEE Trans. Parallel Distrib. Syst.}, vol.~26, no.~4, pp.
  974--983, 2015.

\bibitem{du2018computation}
J.~Du, L.~Zhao, J.~Feng, and X.~Chu, ``Computation offloading and resource
  allocation in mixed fog/cloud computing systems with min-max fairness
  guarantee,'' \emph{IEEE Trans. Commun.}, vol.~66, no.~4, 2018.

\bibitem{guo2018mobile}
H.~Guo, J.~Liu, J.~Zhang, W.~Sun, and N.~Kato, ``Mobile-edge computation
  offloading for ultra-dense iot networks,'' \emph{IEEE Internet of Things
  Journal}, 2018.

\bibitem{zhang2018energy}
J.~Zhang, X.~Hu, Z.~Ning, E.~C.~. Ngai, L.~Zhou, J.~Wei, J.~Cheng, and B.~Hu,
  ``Energy-latency tradeoff for energy-aware offloading in mobile edge
  computing networks,'' \emph{IEEE Internet of Things Journal}, vol.~5, no.~4,
  pp. 2633--2645, 2018.

\bibitem{bi2018computation}
S.~Bi and Y.~J. Zhang, ``Computation rate maximization for wireless powered
  mobile-edge computing with binary computation offloading,'' \emph{IEEE Trans.
  Wireless Commun.}, vol.~17, no.~6, pp. 4177--4190, 2018.

\bibitem{ding2018impact}
Z.~Ding, P.~Fan, and H.~V. Poor, ``Impact of non-orthogonal multiple access on
  the offloading of mobile edge computing,'' \emph{arXiv preprint
  arXiv:1804.06712}, 2018.

\bibitem{zhang2013energy}
W.~Zhang, Y.~Wen, K.~Guan, D.~Kilper, H.~Luo, and D.~O. Wu, ``Energy-optimal
  mobile cloud computing under stochastic wireless channel,'' \emph{IEEE Trans.
  Wireless Commun.}, vol.~12, no.~9, pp. 4569--4581, 2013.

\bibitem{kwak2015dream}
J.~Kwak, Y.~Kim, J.~Lee, and S.~Chong, ``Dream: Dynamic resource and task
  allocation for energy minimization in mobile cloud systems,'' \emph{IEEE J.
  Sel. Areas Commun.}, vol.~33, no.~12, pp. 2510--2523, 2015.

\bibitem{sardellitti2015joint}
S.~Sardellitti, G.~Scutari, and S.~Barbarossa, ``Joint optimization of radio
  and computational resources for multicell mobile-edge computing,'' \emph{IEEE
  Trans. Signal Inf. Process. Over Netw.}, vol.~1, no.~2, pp. 89--103, 2015.

\bibitem{mao2016dynamic}
Y.~Mao, J.~Zhang, and K.~B. Letaief, ``Dynamic computation offloading for
  mobile-edge computing with energy harvesting devices,'' \emph{IEEE J. Sel.
  Areas Commun.}, vol.~34, no.~12, pp. 3590--3605, 2016.

\bibitem{mao2017stochastic}
Y.~Mao, J.~Zhang, S.~Song, and K.~B. Letaief, ``Stochastic joint radio and
  computational resource management for multi-user mobile-edge computing
  systems,'' \emph{IEEE Trans. Wireless Commun.}, vol.~16, no.~9, pp.
  5994--6009, 2017.

\bibitem{lyu2017optimal}
X.~Lyu, W.~Ni, H.~Tian, R.~P. Liu, X.~Wang, G.~B. Giannakis, and A.~Paulraj,
  ``Optimal schedule of mobile edge computing for internet of things using
  partial information,'' \emph{IEEE J. Sel. Areas Commun.}, vol.~35, no.~11,
  pp. 2606--2615, 2017.

\bibitem{chen2018multi}
W.~Chen, D.~Wang, and K.~Li, ``Multi-user multi-task computation offloading in
  green mobile edge cloud computing,'' \emph{IEEE Trans. on Services Comput.},
  2018.

\bibitem{liu2016delay}
J.~Liu, Y.~Mao, J.~Zhang, and K.~B. Letaief, ``Delay-optimal computation task
  scheduling for mobile-edge computing systems,'' in \emph{Proc. IEEE ISIT},
  2016, pp. 1451--1455.

\bibitem{hong2016qoe}
S.-T. Hong and H.~Kim, ``Qoe-aware computation offloading scheduling to capture
  energy-latency tradeoff in mobile clouds,'' in \emph{Proc. IEEE SECON}, 2016,
  pp. 1--9.

\bibitem{xu2017online}
J.~Xu, L.~Chen, and S.~Ren, ``Online learning for offloading and autoscaling in
  energy harvesting mobile edge computing,'' \emph{IEEE Trans. on Cognitive
  Commun. and Netw.}, vol.~3, no.~3, pp. 361--373, 2017.

\bibitem{dinh2018distributed}
T.~Q. Dinh, Q.~D. La, T.~Q. Quek, and H.~Shin, ``Distributed learning for
  computation offloading in mobile edge computing,'' \emph{IEEE Trans.
  Commun.}, 2018.

\bibitem{sutton1998reinforcement}
R.~S. Sutton, A.~G. Barto \emph{et~al.}, \emph{Reinforcement learning: An
  introduction}.\hskip 1em plus 0.5em minus 0.4em\relax MIT press, 1998.

\bibitem{mnih2015human}
V.~Mnih, K.~Kavukcuoglu, D.~Silver, A.~A. Rusu, J.~Veness, M.~G. Bellemare,
  A.~Graves, M.~Riedmiller, A.~K. Fidjeland, G.~Ostrovski \emph{et~al.},
  ``Human-level control through deep reinforcement learning,'' \emph{Nature},
  vol. 518, no. 7540, p. 529, 2015.

\bibitem{alqerm2017energy}
I.~AlQerm and B.~Shihada, ``Energy efficient power allocation in multi-tier 5g
  networks using enhanced online learning,'' \emph{IEEE Trans. Veh. Technol.},
  2017.

\bibitem{he2018green}
X.~He, K.~Wang, H.~Huang, T.~Miyazaki, Y.~Wang, and S.~Guo, ``Green resource
  allocation based on deep reinforcement learning in content-centric iot,''
  \emph{IEEE Trans. Emerging Topics in Comput.}, 2018.

\bibitem{nasir2018deep}
Y.~S. Nasir and D.~Guo, ``Deep reinforcement learning for distributed dynamic
  power allocation in wireless networks,'' \emph{arXiv preprint
  arXiv:1808.00490}, 2018.

\bibitem{chen2018optimized}
X.~Chen, H.~Zhang, C.~Wu, S.~Mao, Y.~Ji, and M.~Bennis, ``Optimized computation
  offloading performance in virtual edge computing systems via deep
  reinforcement learning,'' \emph{arXiv preprint arXiv:1805.06146}, 2018.

\bibitem{huang2018deep}
L.~Huang, S.~Bi, and Y.-J.~A. Zhang, ``Deep reinforcement learning for online
  offloading in wireless powered mobile-edge computing networks,'' \emph{arXiv
  preprint arXiv:1808.01977}, 2018.

\bibitem{min2017learning}
M.~Min, D.~Xu, L.~Xiao, Y.~Tang, and D.~Wu, ``Learning-based computation
  offloading for iot devices with energy harvesting,'' \emph{arXiv preprint
  arXiv:1712.08768}, 2017.

\bibitem{li2018deep}
J.~Li, H.~Gao, T.~Lv, and Y.~Lu, ``Deep reinforcement learning based
  computation offloading and resource allocation for mec,'' in \emph{Proc. IEEE
  WCNC}, 2018, pp. 1--6.

\bibitem{lillicrap2016continuous}
T.~P. Lillicrap, J.~J. Hunt, A.~Pritzel, N.~Heess, T.~Erez, Y.~Tassa,
  D.~Silver, and D.~Wierstra, ``Continuous control with deep reinforcement
  learning,'' in \emph{Proc. International Conference on Learning
  Representations (ICLR)}, 2016.

\bibitem{watkins1992qlearning}
C.~J. C.~H. Watkins and P.~Dayan, ``Q-learning,'' \emph{{Machine Learning}},
  vol.~8, no.~3, pp. 279--292, May 1992.

\bibitem{silver2014deterministic}
D.~Silver, G.~Lever, N.~Heess, T.~Degris, D.~Wierstra, and M.~Riedmiller,
  ``Deterministic policy gradient algorithms,'' in \emph{International
  Conference on Machine Learning}, 2014, pp. 387--395.

\bibitem{arulkumaran2017deep}
K.~Arulkumaran, M.~P. Deisenroth, M.~Brundage, and A.~A. Bharath, ``Deep
  reinforcement learning: A brief survey,'' \emph{IEEE Signal Process. Mag},
  vol.~34, no.~6, pp. 26--38, 2017.

\bibitem{ngo2013energy}
H.~Q. Ngo, E.~G. Larsson, and T.~L. Marzetta, ``Energy and spectral efficiency
  of very large multiuser mimo systems,'' \emph{IEEE Trans. Commun.}, vol.~61,
  no.~4, pp. 1436--1449, 2013.

\bibitem{suraweera2011effect}
H.~A. Suraweera, T.~A. Tsiftsis, G.~K. Karagiannidis, and A.~Nallanathan,
  ``Effect of feedback delay on amplify-and-forward relay networks with
  beamforming,'' \emph{IEEE Trans. Veh. Technol.}, vol.~60, no.~3, pp.
  1265--1271, 2011.

\bibitem{abramowitz1972handbook}
M.~Abramowitz, I.~A. Stegun \emph{et~al.}, \emph{Handbook of mathematical
  functions: with formulas, graphs, and mathematical tables}.\hskip 1em plus
  0.5em minus 0.4em\relax Dover publications New York, 1972, vol.~55.

\bibitem{miettinen2010energy}
A.~P. Miettinen and J.~K. Nurminen, ``Energy efficiency of mobile clients in
  cloud computing.'' \emph{HotCloud}, vol.~10, pp. 4--4, 2010.

\bibitem{burd1996processor}
T.~D. Burd and R.~W. Brodersen, ``Processor design for portable systems,''
  \emph{Journal of VLSI signal processing systems for signal, image and video
  technology}, vol.~13, no. 2-3, pp. 203--221, 1996.

\bibitem{tse2005fundamentals}
D.~Tse and P.~Viswanath, \emph{Fundamentals of wireless communication}.\hskip
  1em plus 0.5em minus 0.4em\relax Cambridge university press, 2005.

\bibitem{shortle2018fundamentals}
J.~F. Shortle, J.~M. Thompson, D.~Gross, and C.~M. Harris, \emph{Fundamentals
  of queueing theory}.\hskip 1em plus 0.5em minus 0.4em\relax John Wiley \&
  Sons, 2018, vol. 399.

\bibitem{adelman2008relaxations}
D.~Adelman and A.~J. Mersereau, ``Relaxations of weakly coupled stochastic
  dynamic programs,'' \emph{Operations Research}, vol.~56, no.~3, pp. 712--727,
  2008.

\bibitem{tensorflow2015-whitepaper}
\BIBentryALTinterwordspacing
A.~Mart\'{\i}n, A.~Ashish \emph{et~al.}, ``{TensorFlow}: Large-scale machine
  learning on heterogeneous systems,'' 2015, software available from
  tensorflow.org. [Online]. Available: \url{https://www.tensorflow.org/}
\BIBentrySTDinterwordspacing

\bibitem{kingma2014adam}
D.~P. Kingma and J.~Ba, ``Adam: A method for stochastic optimization,''
  \emph{arXiv preprint arXiv:1412.6980}, 2014.

\bibitem{uhlenbeck1930theory}
G.~E. Uhlenbeck and L.~S. Ornstein, ``On the theory of the brownian motion,''
  \emph{Physical review}, vol.~36, no.~5, p. 823, 1930.

\end{thebibliography}
\end{document}